\documentclass[10pt,twocolumn,letterpaper]{article}

\usepackage{iccv}
\usepackage{times}
\usepackage{epsfig}
\usepackage{graphicx}
\usepackage{amsmath}
\usepackage{amssymb}
\usepackage{lipsum}
\usepackage{subcaption}
\usepackage{booktabs}
\usepackage{colortbl}
\usepackage[export]{adjustbox}
\usepackage[utf8]{inputenc}
\usepackage{authblk}

\usepackage[pagebackref=true,breaklinks=true,letterpaper=true,colorlinks,bookmarks=false]{hyperref}
\usepackage{color}
\iccvfinalcopy 

\def\iccvPaperID{318} 

\definecolor{cwblue1}{rgb}{0.27,0.427,0.623}
\definecolor{cwblue2}{rgb}{0.286,0.454,0.658}
\definecolor{cwblue3}{rgb}{0.733,0.811,0.905}

\newcommand{\bh}{\mbox{\boldmath$h$}}

\newcommand{\bp}{\mbox{\boldmath$p$}}

\newcommand{\bs}{\mbox{\boldmath$s$}}
\newcommand{\bu}{\mbox{\boldmath$u$}}
\newcommand{\bv}{\mbox{\boldmath$v$}}

\newcommand{\bx}{\mbox{\boldmath$x$}}

\newcommand{\bI}{\mbox{\boldmath$I$}}

\newcommand{\bP}{\mbox{\boldmath$P$}}

\newcommand{\bU}{\mbox{\boldmath$U$}}
\newcommand{\bV}{\mbox{\boldmath$V$}}

\newcommand{\bX}{\mbox{\boldmath$X$}}

\newcommand{\myparagraph}[1]{\textbf{#1} --- }  

\ificcvfinal\fi
\begin{document}

\title{Pose-conditioned Spatio-Temporal Attention for Human Action Recognition}

\author[1]{Fabien Baradel}
\author[1]{Christian Wolf}
\author[2]{Julien Mille}
\affil[1]{Univ Lyon, INSA-Lyon, CNRS, LIRIS, F-69621, Villeurbanne, France}
\affil[2]{Laboratoire d'Informatique de l'Université de Tours (EA 6300), INSA Centre Val de Loire, 41034 Blois, France}
\affil[ ]{\small { \{fabien.baradel, christian.wolf\}@liris.cnrs.fr, julien.mille@insa-cvl.fr}}

\maketitle

\begin{abstract}
    \noindent
	We address human action recognition from multi-modal video data involving articulated pose and RGB frames and propose a two-stream approach. The pose stream is processed with a convolutional model taking as input a 3D tensor holding data from a sub-sequence. A specific joint ordering, which respects the topology of the human body, ensures that different convolutional layers correspond to meaningful levels of abstraction.
	
	The raw RGB stream is handled by a spatio-temporal soft-attention mechanism conditioned on features from the pose network. An LSTM network receives input from a set of image locations at each instant. A trainable glimpse sensor extracts features on a set of predefined locations specified by the pose stream, namely the 4 hands of the two people involved in the activity. Appearance features give important cues on hand motion and on objects held in each hand. We show that it is of high interest to shift the attention to different hands at different time steps depending on the activity itself. 
    Finally a temporal attention mechanism learns how to fuse LSTM features over time.
    
	We evaluate the method on 3 datasets. State-of-the-art results are achieved on the largest dataset for human activity recognition, namely NTU-RGB+D, as well as on the SBU Kinect Interaction dataset. Performance close to state-of-the-art is achieved on the smaller MSR Daily Activity 3D dataset.
	
\end{abstract}

\section{Introduction}

\begin{figure}[t]
    \centering
        \includegraphics[width=8cm]{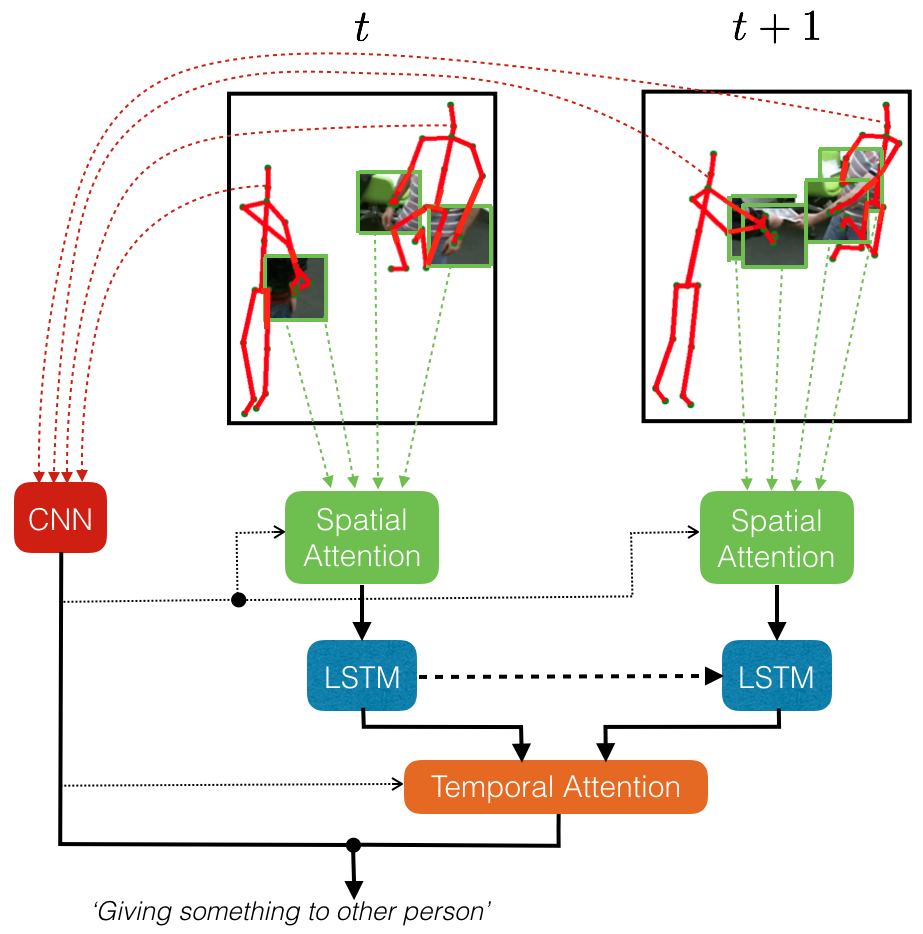}
    \caption{We recognize human activities fusing a model trained on pose sub-sequences and a spatio-temporal attention model on RGB video conditioned on pose features.
    }
    \label{fig:teaser}
\end{figure}

\noindent
Human activity recognition is a field with many applications ranging from video surveillance, HCI, robotics, to automated driving and others. Consumer depth cameras are currently dominating the field for indoor applications with close ranges, as they allow to estimate articulated poses easily. We address similar settings, namely activity recognition problems where articulated pose is available. As complementary information we also use the RGB stream, which provides rich contextual cues on human activities, for instance on the objects held or interacted with. 

Recognizing human actions accurately remains a challenging task, compared to other problems in computer vision and machine learning. We argue that this is in part due to the lack of large datasets. While large scale datasets have been available for a while for object recognition (ILSVRC~\cite{Russakovsky2015}) and for general video classification (Sports-1M \cite{KarpathyCVPR14} and lately Youtube8M \cite{Youtube8M2016}), the more time-consuming acquisition process for videos showing close range human activities limited datasets of this type to several hundreds or a few thousand videos. As a consequence, the best performing methods on this kind of datasets are either based on handcrafted features or suspected to overfit on the small datasets after years the community spent on tuning methods. The recent introduction of datasets like NTU-RGB-D \cite{Shahroudy2016} (${\sim}$ 57 000 videos) will hopefully lead to better automatically learned representations.

One of the challenges is the high amount of information in videos.
Downsampling is an obvious choice, but using the full resolution at certain positions may help extracting important cues on small or far away objects (or people). In this regard, models of visual attention \cite{Mnih_NIPS2014,ChoBengioMM2015,Sharma2016a} (see section \ref{sec:relatedworks} for a full discussion) have drawn considerable interest recently. Capable of focusing their attention to specific important points, parameters are not wasted on input which is considered of low relevance to the task at hand.

We propose a method for human activity recognition, which addresses this problem by fusing articulated pose and raw RGB input in a novel way. In our approach, pose has three complementary roles: i) it is used as an input stream in its own right, providing important cues for the discrimination of activity classes; ii) raw pose (joints) serves as an input for the model handling the RGB stream, selecting positions where glimpses are taken in the image; iii) features learned on pose serve as an input to the soft-attention mechanism, which weights each glimpse output according to an estimated importance w.r.t. the task at hand, in contrast to unconstrained soft-attention on RGB video \cite{Sharma2016a}.

The RGB stream model is recurrent (an LSTM), whereas our pose representation is learned using a convolutional neural network taking as input a sub-sequence of the video. The benefits are twofold: a pose representation over a large temporal range allows the attention model to assign an estimated importance for each glimpse point and each time instant taking into account knowledge of this temporal range. As an example, the pose stream might indicate that the hand of one person moves into the direction of a different person, which still leaves several possible choices for the activity class. These choices might require attention to be moved to this hand at a specific instant  to verify what kind of object is held, which itself may help to discriminate activities.

The contributions of our work are as follows:
\begin{itemize}

	\item We propose a way to encode articulated pose data over time into 3D tensors which can be fed to CNNs as an alternative to recurrent neural networks. We propose a particular joint ordering which preserves neighborhood relationships between the joints in the body.

	\item We propose a spatial attention mechanism on RGB videos which is conditioned on pose features from the full sub-sequence. 

    \item We propose a temporal attention mechanism which learns how to pool features output from the recurrent (LSTM) network over time in an adaptive way.

	\item As an additional contribution, we experimentally show that knowledge transfer from a large activity dataset like NTU (57000 activities) to smaller datasets like MSR Daily Activitiy 3D (300 videos) is possible. Up to our knowledge, this ImageNet-style transfer has not been attempted on human activities.
\end{itemize}

Animated video can be found on the project page\footnote{\url{https://fabienbaradel.github.io/pose_rgb_attention_human_action}}.

\section{Related Work}
\label{sec:relatedworks}
\noindent
\myparagraph{Activities, gestures and multimodal data}
Recent gesture/action recognition methods dealing with several modalities typically process 2D+T RGB and/or depth data as 3D. Sequences of frames are stacked into volumes and fed into convolutional layers at first stages~\cite{Baccouche2011,Ji_PAMI2013, MolchanovYangCVPR2016, NeverovaWolfTaylorNeboutPAMI2016, WuPigouPAMI2016}. When additional pose data is available, the 3D joint positions are typically fed into a separate network. Preprocessing pose is reported to improve performance in some situations, e.g. augmenting coordinates with velocities and acceleration~\cite{DBLP:conf/iccv/ZanfirLS13}. Pose normalization (bone lengths and view point normalization) has been reported to help in certain situations \cite{NeverovaWolfTaylorNeboutPAMI2016}.
Fusing pose and raw video modalities is traditionally done as late fusion~\cite{MolchanovYangCVPR2016}, or early through fusion layers~\cite{WuPigouPAMI2016}. In \cite{LiNeverovaWolfTaylor2017}, fusion strategies are learned together with model parameters with by stochastic regularization.







  
\myparagraph{Recurrent architectures for action recognition}
Most recent activity recognition methods 
are based on recurrent neural networks in some form. In the variant Long Short-Term Memory (LSTM) ~\cite{Hochreiter1997}, a gating mechanism over an internal memory cell learns long-term and short-term dependencies in the sequential input data. Part-aware LSTMs ~\cite{Shahroudy2016} separate the memory cell into part-based sub-cells and let the network learn long-term representations individually for each part, fusing the parts for output. 
Similarly, Du {\it et al}~\cite{Du_CVPR2015} use bi-directional LSTM layers which fit anatomical hierarchy. Skeletons are split into anatomically-relevant parts (legs, arms, torso, {\it etc}), so that each subnetwork in the first layers gets specialized on one part. Features are progressively merged as they pass through layers. 

Multi-dimensional LSTMs \cite{Graves2DLSTM2009} are models with multiple recurrences from different dimensions. Originally introduced for images, they also have been applied to activity recognition from pose sequences~\cite{Liu2016}. One dimension is time, the second is a topological traversal of the joints in a bidirectional depth-first search, which preserves the neighborhood relationships in the graph. Our solution for pose features a similar joint traversal. However, our pose network is convolutional and not recurrent (whereas our RGB network is recurrent).




\myparagraph{Attention mechanisms} 
Human perception focuses selectively on parts of the scene to
acquire information at specific places and times. In machine learning, this kind of processes is referred to as attention mechanism, and has drawn increasing interest when dealing with languages, images and other data. Integrating attention can potentially lead to improved overall accuracy, as the system can focus on parts of the data, which are most relevant to the task.
 
In computer vision, visual attention mechanisms date as far back as the work of Itti {\it et al} for object detection~\cite{Itti_PAMI1998}. Early models were highly related to saliency maps, i.e. pixelwise weighting of image parts that locally stand out, no learning was involved. Larochelle and Hinton~\cite{Larochelle_NIPS2010} pioneered the incorporation of attention into a learning architecture by coupling Restricted Boltzmann Machines with a foveal representation.

More recently, attention mechanisms were gradually categorized into two classes. \emph{Hard attention} takes hard decisions when choosing parts of the input data. This leads to stochastic algorithms, which cannot be easily learned through gradient descent and back-propagation. In a seminal paper, Mnih {\it et al}~\cite{Mnih_NIPS2014} proposed visual hard-attention for image classification built around a recurrent network, which implements the policy of a virtual agent. A reinforcement learning problem is thus solved during learning~\cite{Williams1992}. The model selects the next location to focus on, based on past information. Ba et al ~\cite{Ba-attention-2015} improved the approach to tackle multiple object recognition. In \cite{Kuen_CVPR2015}, a hard attention model generates saliency maps. Yeung {\it et al}~\cite{Yeung_CVPR2016} use hard-attention for action detection with a model, which decides both which frame to observe next as well as when to emit an action prediction. 

On the other hand, \emph{soft attention} takes the entire input into account, weighting each part of the observations dynamically. The objective function is usually differentiable, making gradient-based optimization possible. Soft attention was used for various applications such as neural machine translation~\cite{Bahdanau_ICLR2015, Kim_ICLR2017} or image captioning~\cite{Xu_ICML2015}.
Recently, soft attention was proposed for image \cite{ChoBengioMM2015} and video understanding ~\cite{Sharma2016a,Song2016,yeung2015every}, with spatial, temporal and spatio-temporal variants. Sharma {\it et al}~\cite{Sharma2016a} proposed a recurrent mechanism for action recognition from RGB data, which integrates convolutional features from different parts of a space-time volume. Yeung et al. report a temporal recurrent attention model for dense labelling of videos~\cite{yeung2015every}. At each time step, multiple input frames are integrated and soft predictions are generated for multiple frames.
Bazzani {\it et al} \cite{Bazzani_ICLR2017} learn spatial saliency maps represented by mixtures of Gaussians, whose parameters are included into the internal state of a LSTM network. 
Saliency maps are then used to smoothly select areas with relevant human motion.
Song {\it et al} \cite{Song2016} propose separate spatial and temporal attention networks for action recognition from pose. At each frame, the spatial attention model gives more importance to the joints most relevant to the current action, whereas the temporal model selects frames.


Up to our knowledge, no attention model has yet taken advantage of both articulated pose and RGB data simultaneously.
Our method has slight similarities with hard attention in that hard choices are taken on locations in each frame. However, these choices are not learned, they depend on pose. On the other hand, we learn a soft-attention mechanism, which dynamically weights features from several locations. The mechanism is conditional on pose, which allows it to steer its focus depending on motion.


\begin{figure*} \centering
    \centering
        \includegraphics[width=14cm]{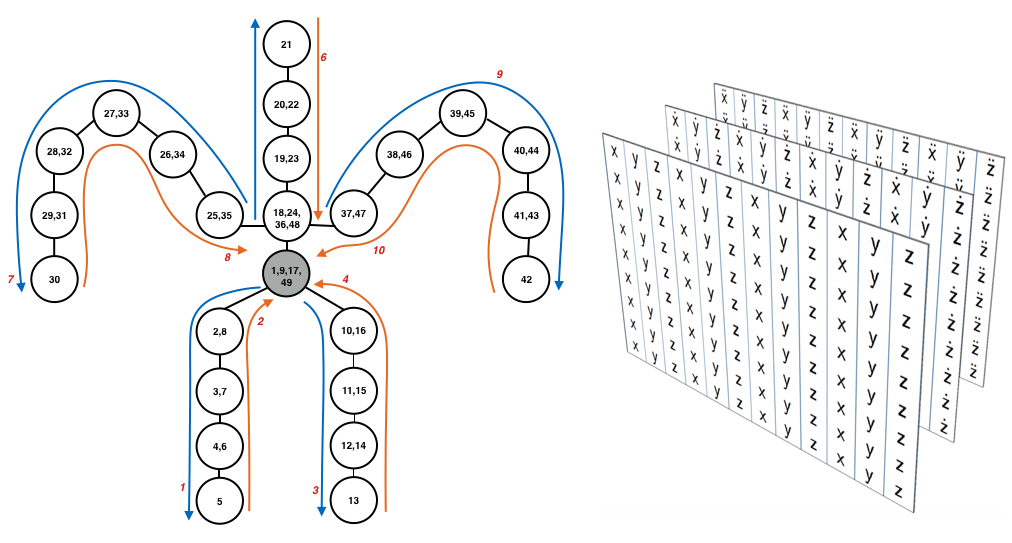} \\
        (a) \hspace{5cm} (b)
        \caption{\label{fig:toposkel} (a) the topological ordering of joints (similar to \cite{Liu2016}): blue arrows visit joints for the first time and orange arrows go back to the ``middle spine''. (b) the ordering is reproduced in the matrix input to the pose learner)}
\end{figure*}

\section{Proposed Model}
\label{sec:proposed-model}

\noindent
A single or multi-person activity is described by a sequence of two modalities: the set of RGB input images $\bI{=}\{ I_t \}$, and the set of articulated human poses $\bx{=}\{ \bx_t \}$. We do not use raw depth data in our method, although the extension would be straightforward. Both signals are indexed by time $t$. Poses $\bx_t$ are defined by 3D coordinates of joints, for instance 
delivered by the middleware of a depth camera.
The sheer amount of data per input sequence makes it difficult to train a classical (convolutional or recurrent) model directly on the sequence $\{\bI_0, \dots, \bI_T, \ \bx_0, \dots, \bx_T\}$ of inputs to predict activity classes $y$.
We propose a two-stream model, which classifies activity sequences by extracting features from articulated human poses and RGB frames.

\subsection{Convolutional pose features}
\label{subsec:skeletonnet}

\noindent
At each time step $t$, a subject is represented by the 3D coordinates of its $K$ body joints. In our case we restrict our application to activities involving one or two people and their interactions. The goal is to extract features which model i) the temporal behavior of the pose(s) and ii) correlations between different joints. An attention mechanism on poses could be an option, similar to \cite{Song2016}. We argue that the available pose information is sufficiently compact to learn a global representation and show that this is efficient. However, we also argue for the need to find a hierarchical representation which respects the spatio-temporal relationships of the data. In the particular case of pose data, joints also have strong neighborhood relationships in the human body.

In the lines of \cite{Liu2016}, we define a topological ordering of the joints in a human body as a connected cyclic path over joints (see figure \ref{fig:toposkel}a). The path itself is not Hamiltonian as each node can be visited multiple times: once during a forward pass over a limb, and once during a backward pass over the limb back to the joint it is attached to. The double entries in the path are important, since they ensure that the path preserves neighborhood relationships.

In \cite{Liu2016}, a similar path is used to define an order in a multi-dimensional LSTM network. In contrast, we propose a convolutional model which takes three-dimensional inputs (tensors) calculated by concatenating pose vectors over time. In particular, input tensors $\bX$ are defined as 
$\bX {=} \{ \bX_{t,j,k} \}$, where $t$ is the time index, $j$ is the joint \& coordinate index, and $k$ is a feature index (see figure \ref{fig:toposkel}b): each line corresponds to a time instant; the first three columns correspond to the $x$, $y$ and $z$ coordinates of the first joint, followed by the $x$, $y$ and $z$ coordinate of the second joint, which is a neighbor of the first etc. The first channel corresponds to raw coordinates, the second channel corresponds to first derivates of coordinates (velocities), the third channel to second derivates (accelerations). Poses of two people are stacked into a single tensor along the second dimension. This choice of tensor organization will be justified further below.

We learn a pose network $f_{sk}$ with parameters $\theta_{sk}$ on this input, resulting in a pose feature representation $\bs$:
\begin{equation}
\bs = f_{sk} (\bX, \theta_{sk})
\end{equation}
Here and in the rest of the paper, subscripts of mappings $f$ and their parameters $\theta$ choose a specific mapping, they are not indices. Subscripts of variables and tensors are indices.

$f_{sk}$ is implemented as a convolutional neural network alternating convolutions and max-pooling. Combined with the topological ordering of the columns of the input tensor, this leads to a specific hierarchical representation of the feature maps. The first layer of convolutions will extract features from the correlations between coordinates, mostly of the same joints (or neighboring joints). Subsequent convolutions will extract features between neighboring joints, and even higher layers in the network correspond to extractions of features which are further away in the human body, in the sense of path lengths in the graph. The last layers correspond to features extracted between the two different poses corresponding to two different people.

One design choice of this representation is to stack different coordinates $(x,y,z)$ of the same joint into subsequent columns of the tensor, opposed to the alternative of distributing them over different channels. This ensures, that the first layer calculates features on different coordinates. Experiments have confirmed the interest of this choice. The double entries in the input tensor $\bX$ artificially increase its size, as some joints are represented multiple times. However, this cost is compensated by the fact that the early convolutional layers extract features on joint pairs which are neighbors in the graph (in the human body). 

\subsection{Spatial Attention on RGB videos}

\noindent
The sequence of RGB input images $\{\bI_t\}$ is arguably not compact enough to easily extract an efficient global representation with a feed-forward neural network. We opt for a recurrent solution, where, at each time instant, a glimpse on the seen input is selected using an attention mechanism.

In some aspects similar to \cite{Mnih_NIPS2014}, we define a trainable bandwith limited sensor. However, in contrast to \cite{Mnih_NIPS2014}, our attention process is conditional to the pose input $\bx_t$, thus limited to a set of $N$ discrete attention points. In our experiments, we selected $N{=}4$ attention points, which are the 4 hand joints of the two people involved in the interaction. The goal is to extract additional information about hand shape and about manipulated objects. A large number of activities such as \emph{Reading}, \emph{Writing}, \emph{Eating}, \emph{Drinking} are similar in motion but can be highly correlated to manipulated objects. As the glimpse location is not output by the network, this results in a differentiable soft-attention mechanism, which can be trained by gradient descent.

\begin{figure}[t!] \centering
        \includegraphics[width=8cm]{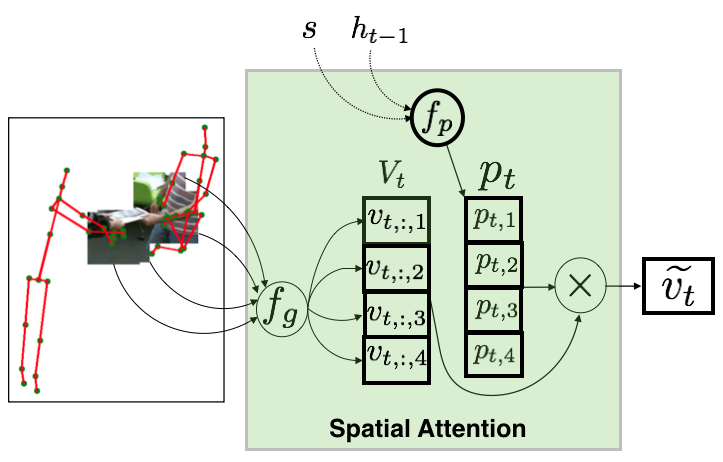}
        \caption{\label{fig:sattention} The spatial attention mechanism}
\end{figure}

The glimpse representation for a given attention point $i$ is a convolutional network $f_g$ with parameters $\theta_g$, taking as inputs a crop taken from image $I_t$ at the position of joint $i$ from the set $\bx_t$:
\begin{equation}
\bv_{t,:,i} = f_g (\textrm{crop}(I_t,\bx_t,i),\theta_g)  {\small \quad \quad i{=}\{1,\dots N\} }
\end{equation}
Here, $\bv_{t,:,i}$ is a (column) feature vector for time $t$ and hand $i$. For a given time $t$, we stack the vectors into a matrix $\bV_t {=}\{ \bv_{t,j,i} \}$, where $i$ is the index over hand joints and $j$ is the index over features. $\bV_t$ is a matrix (a 2D tensor), since t is fixed for a given instant.
 
A recurrent model receives inputs from the glimpse sensor sequentially and models the information from the seen sequence with a componential hidden state $\bh_t$:
\begin{equation}
\bh_t = f_h (\bh_{t-1}, \tilde{\bv_t}, \theta_h)
\label{eq:lstm}
\end{equation}
We chose a fully gated LSTM model including input, forget and output gates and a cell state. To keep the notation simple, we omitted the gates and the cell state from the equations. The input to the LSTM network is the context vector $\tilde{\bv}_t$, defined further below, which corresponds to an integration of the different attention points (hands) in $\bV_t$. 

An obvious choice of integration are simple functions like sums and concatenations. While the former tends to squash feature dynamics by pooling strong feature activations in one hand with average or low activations in other hands, the latter leads to high capacity models with low generalization. The soft-attention mechanism dynamically weighs the integration process through a distribution $\bp_t$, determining how much attention hand $i$ needs with a calculated weight $\bp_{t,i}$. 
In contrast to unconstrained soft-attention mechanisms on RGB video \cite{Sharma2016a}, our attention distributions not only depends on the LSTM state $\bh_t$, but also on the pose features $\bs$ extracted from the sub-sequence, through a learned mapping with parameters $\theta_p$:
\begin{equation}
\bp_t = f_p (\bh_{t}, \bs, \theta_p)
\label{eq:sattention}
\end{equation}
Attention distribution $\bp_t$ and features $\bV_t$ are integrated through a linear combination as
\begin{equation}
\tilde{\bv}_t = \bV_{t} \bp_{t} \ ,
\end{equation}
which is input to the LSTM network at time $t$ (see eq. (\ref{eq:lstm})). The conditioning on the pose features in \ref{eq:sattention} is important, as it provides valuable context derived from motion. Note that the recurrent model itself (eq. (\ref{eq:lstm})) is not conditional \cite{MikolovZweig2012}, this would significantly increase the amount of parameters.

\begin{figure}[t!] \centering
        \includegraphics[width=9cm, height=12cm]{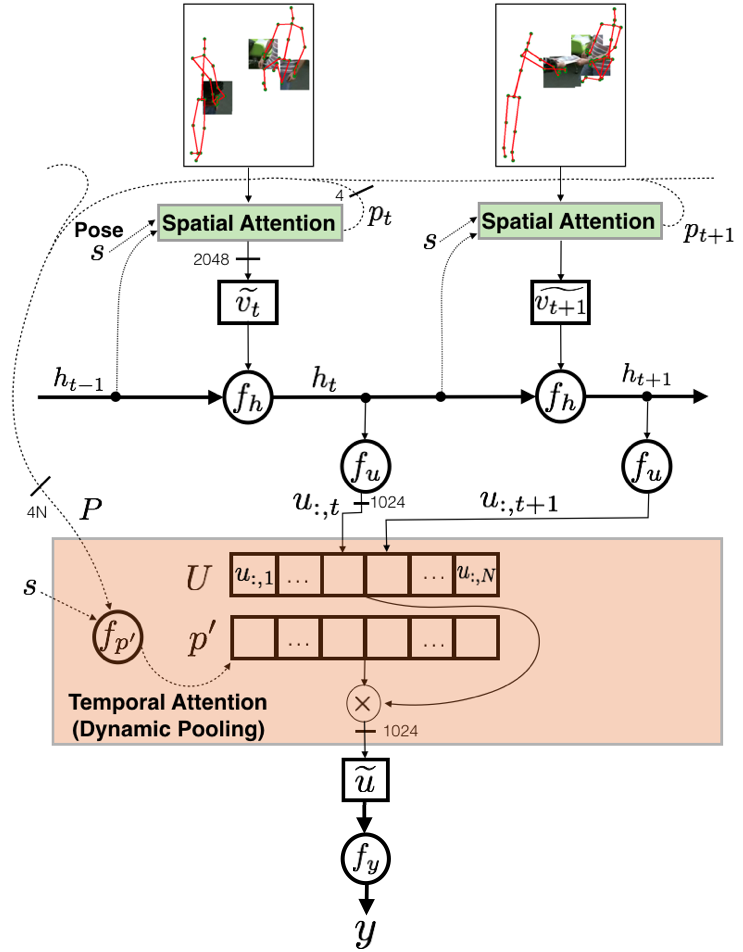}
        \caption{\label{fig:tattention} The full recurrent model for RGB data (gates and memory cell are not shown). Pose $\bs$ is input to the attention mechanism. The spatial mechanism is detailed in figure \ref{fig:sattention}.}
\end{figure}

\begin{figure*}[t] \centering
    \includegraphics[width=12cm]{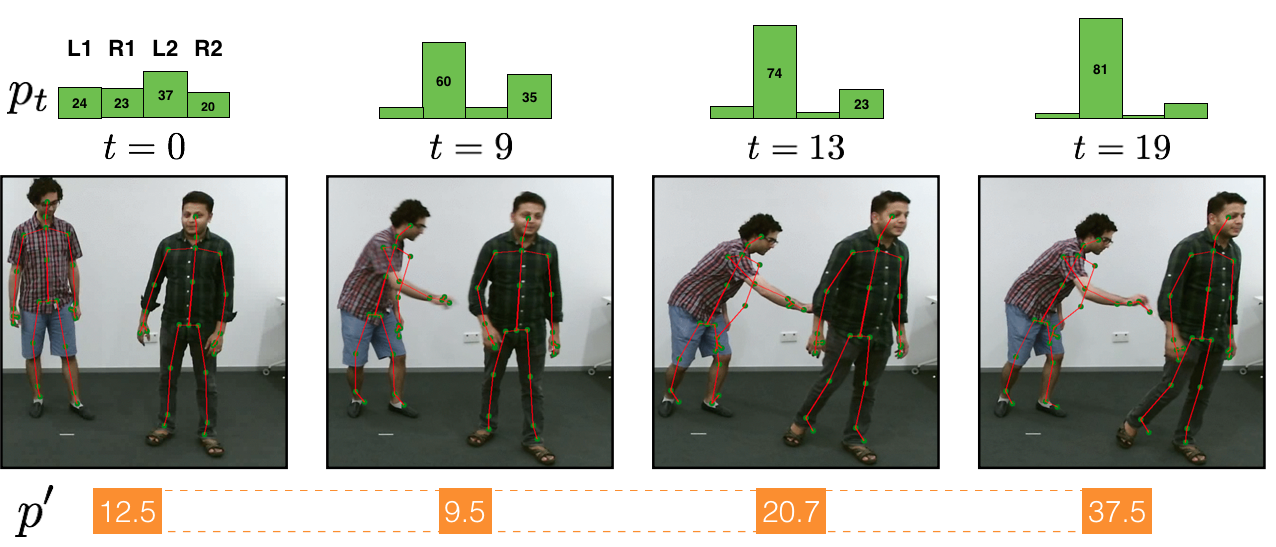}
    \caption{\label{fig:attentionexample} Spatial attention over time: putting an object into the pocket of someone will make the attention shift to this hand.}
\end{figure*}

\subsection{Temporal Attention}

\noindent
Recurrent models can provide predictions for each time step $t$. Most current work in sequence classification proceeds by temporal pooling of these predictions, e.g. through a sum or average \cite{Sharma2016a}. 
We show that it can be important to perform this pooling in an adaptive way. In recent work on dense activity labelling, temporal attention for dynamical pooling of LSTM logits has been proposed \cite{yeung2015every}. In contrast, we perform temporal pooling directly on feature vector level. In particular, at each instant $t$, features are calculated by a learned mapping given the current hidden state:
\begin{equation}
\bu_{:,t} = f_u (\bh_t, \theta_u)
\end{equation}
The features for all instants $t$ of the sub-sequence are stacked into a matrix $\bU{=}\{\bu_{j,t}\}$, where $j$ is the index over the feature dimension. A temporal attention distribution $\bp'$ is predicted through a learned mapping. To be efficient, this mapping should have seen the full sub-sequence before giving a prediction for an instant t, as giving a low weight to features at the beginning of a sequence might be caused by the need to give higher weights to features at the end. In the context of sequence-to-sequence alignment, this has been addressed with bi-directional recurrent networks \cite{Bahdanau2014}. To keep the model simple, we benefit from the fact that (sub) sequences are of fixed length and that spatial attention information is already available. We conjecture that (combined with pose) the spatial attention distributions $\bp_t$ over time t are a good indicator for temporal attention, and stack them into a single vector $\bP$, input into the network predicting temporal attention:
\begin{equation}
\bp' = f'_p (\bP, \bs, \theta'_p)
\end{equation}
This attention is used as weight for adaptive temporal pooling of the features $\bU$, i.e. 
$
\tilde{\bu} = \bU \bp'
$.

\begin{figure*}[t] \centering
	\includegraphics[width=12cm]{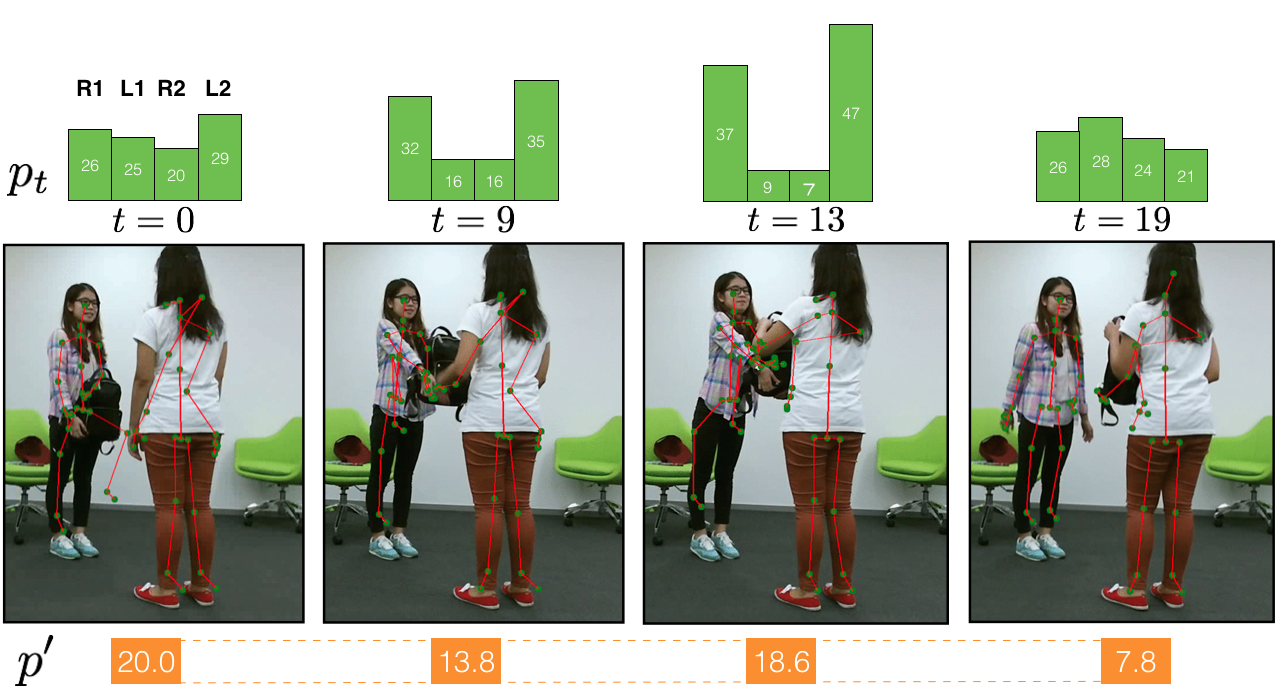}
	\caption{\label{fig:attentionexample2} Spatial and temporal attention over time: giving something to other person will make the attention shift to the active hands in the action.}
\end{figure*}

\subsection{Stream fusion}

\noindent
Each stream, pose and RGB, leads to its own set of features, with the particularity that pose features $\bs$ are input to the attention mechanism for the RGB stream. Each representation is classified with its own set of parameters. We fuse both streams on logit level. More sophisticated techniques, which learn fusion \cite{NeverovaWolfTaylorNeboutPAMI2016}, do not seem to be necessary. 

\section{Network architectures and Training}
\label{sec:archtrain}

\noindent
\myparagraph{Architectures}
The pose network $f_{sk}$ consists of 3 convolutional layers of respective sizes $8{\times}3$, $8{\times}3$, $5{\times}75$. Inputs are of size $20{\times}300{\times}3$ and feature maps are, respectively, $10{\times}150$, $5{\times}75$ and $1{\times}1{\times}1024$. Max pooling is employed after each convolutional layer, activations are ReLU.

The glimpse senor $f_g$ is implemented as an Inception V3 network \cite{Szegedy2016}. Each vector $\bv_{t,:,i}$ corresponds to the last layer before output and is of size 2048.
The LSTM network $f_h$ has a single recurrent layer with 1024 units. The spatial attention network $f_p$ is an MLP with a single hidden layer of 256 units and sigmoid activation. The temporal attention network $f'_p$ is an MLP with a single hidden layer of 512 units and sigmoid activation. The feature extractor $f_u$ is a single linear layer with ReLU activation.
The output layers of both stream representations are linear layers followed by softmax activation. 
The full model (without glimpse sensor $f_g$) has 38 millions trainable parameters.

\myparagraph{Training}
All classification outputs are softmax activated and trained with cross-entropy loss. The glimpse sensor $f_g$ is 
trained on the ILSVRC 2012 data~\cite{Russakovsky2015}. The pose learner is trained discriminatively with an additional linear+softmax layer to predict action classes. The RGB stream model is trained with pose parameters $\theta_{sk}$ and glimpse parameters $\theta_g$ frozen. End-to-end training the model did not result into better performance.



\section{Experiments}
\label{sec:experiments}

\noindent
The proposed method has been evaluated on three datasets: NTU RGB+D, MSR Daily Activity 3D and SBU Kinect Interaction. We extensively tested on NTU and we shows two transfer experiments on the smaller datasets SBU and MSR.

\myparagraph{NTU RGB+D Dataset (NTU)~\cite{Shahroudy2016}} 
The largest dataset for human activity recognition has been acquired with a Kinect v2 sensor and contains more than 56K videos and 4 millions frames with 60 different activities including individual activities, interactions between 2 people and health related events. The actions have been performed by 40 subjects and with 80 viewpoints.
We follow the cross-subject and cross-view split protocol from~\cite{Shahroudy2016}.

\myparagraph{MSR Daily Activity3D Dataset (MSR)~\cite{conf/cvpr/WangLWY12}}
This dataset is among the most challenging benchmarks due to a high level of intra-class variation. It consists of 320 videos shot with a Kinect v1 sensor. 16 daily activities are performed twice each by 10 subjects from a single viewpoint. 
Following ~\cite{conf/cvpr/WangLWY12}, we use videos from subject 1, 3, 5, 7 and 9 for training, and the remaining ones for testing.

\myparagraph{SBU Kinect Interaction Dataset (SBU)~\cite{kiwon_hau3d12}}
This interaction dataset features two subjects with in total 282 sequences (6822 frames) and 8 mutual activity classes shot with a Kinect v1 sensor. 
We follow the standard experimental protocol \cite{kiwon_hau3d12}, which consists in 5-fold cross validation.

The MSR and SBU datasets are extremely challenging for methods performing representation learning, as only few videos are available for training (160 and 225, respectively).

\begin{table}
	\begin{center}
		\begin{tabular}{cccccc}
			\arrayrulecolor{cwblue1} \toprule
			Methods                                  & {\footnotesize Pose} &  {\footnotesize RGB} & CS & CV & Avg \\ 
            \arrayrulecolor{cwblue1} \toprule
			Lie Group \cite{Vemulapalli_2014_CVPR}     & X & -& 50.1          & 52.8       & 51.5    \\ 
			Skeleton Quads \cite{Evangelidis-ICPR-2014} & X & -&38.6          & 41.4       & 40.0    \\ 
			Dynamic Skeletons \cite{Hu_CVPR2015}      & X & -&60.2          & 65.2       & 62.7    \\ 
			HBRNN \cite{Du_CVPR2015}     & X & -&59.1          & 64.0       & 61.6    \\ 
			Deep LSTM \cite{Shahroudy2016}        & X & -&60.7          & 67.3       & 64.0    \\ 
			Part-aware LSTM \cite{Shahroudy2016}  & X & -&62.9          & 70.3       & 66.6    \\ 
			ST-LSTM + TrustG.  \cite{Liu2016}    & X & -&69.2          & 77.7       & 73.5    \\ 
			STA-LSTM   \cite{Song2016}         & X & -&73.4          & 81.2       & 77.2    \\ 
			JTM   \cite{Wang}          & X & -& 76.3   & 81.1   &  78.7 \\ \hline
            DSSCA - SSLM \cite{Shahroudy20162} & X & X & 74.9 & - & -  \\ \hline
			\textbf{Ours (pose only)}    & X & -& \textbf{77.1}        & \textbf{84.5}     & \textbf{80.8}   \\ 
			\textbf{Ours (RGB only)}    & - & X & \textbf{75.6}        & \textbf{80.5}     & \textbf{78.1}   \\ 
            \textbf{Ours (pose +RGB)}    & X & X& \textbf{84.8}       &   \textbf{90.6} & \textbf{87.7}   \\
            \arrayrulecolor{cwblue1} \bottomrule
		\end{tabular}
	\end{center}
	\caption{Results on the NTU RGB+D dataset with Cross-Subject (CS) and Cross-View (CV) settings (accuracies in \%).}
	\label{NTU}
\end{table}

\begin{table}
	\begin{center}
		\begin{tabular}{ccccc}
			\hline
			\arrayrulecolor{cwblue1} \toprule
			Methods   & {\footnotesize Pose} & {\footnotesize RGB} & {\footnotesize Depth} & Acc. \\ 
			\arrayrulecolor{cwblue1} \toprule
			Raw skeleton \cite{kiwon_hau3d12}   &X&-&- & 49.7          \\ 
			Joint feature \cite{kiwon_hau3d12}    &X&-&- & 80.3            \\ 
			Raw skeleton \cite{Ji2014}   &X&-&- & 79.4           \\ 
			Joint feature \cite{Ji2014}     &X&-&- &86.9            \\ 
			Co-occurence RNN  \cite{ZhuLXZLSX16}&X&-&- & 90.4            \\ 
			STA-LSTM   \cite{Song2016}   &X&-&- & 91.5        \\ 
			ST-LSTM + Trust Gate  \cite{Liu2016}         &X&-&- & 93.3           \\    
			DSPM  \cite{LinWZWLZ15}         &-&X&X & 93.4           \\    
			\arrayrulecolor{cwblue1}   \midrule
			\textbf{Ours (Pose only)}     &X&-&- &  90.5      \\ 
			\textbf{Ours (RGB only)}      &-&X&- &  72.0     \\  
			\textbf{Ours (Pose + RGB)}      &X&X&- &  94.1    \\  
			\arrayrulecolor{cwblue1} \bottomrule
		\end{tabular}
	\end{center}
	\caption{Results on SBU Kinect Interaction dataset (accuracies in \%)}
	\label{SBU}
\end{table}

\begin{table}
	\begin{center}
		\begin{tabular}{ccccc}
			\hline
			\arrayrulecolor{cwblue1} \toprule
			Methods   & {\footnotesize Pose} & {\footnotesize RGB} & {\footnotesize Depth} & Acc. \\ 
			\arrayrulecolor{cwblue1} \toprule
			Action Ensemble \cite{conf/cvpr/WangLWY12}    &X&-&- & 68.0           \\ 
			Efficient Pose-Based \cite{DBLP:conf/accv/EweiwiCBG14}     &X&-&- & 73.1            \\ 
			Moving Pose  \cite{DBLP:conf/iccv/ZanfirLS13}  &X&-&- & 73.8            \\ 
			Moving Poselets \cite{conf/iccvw/TaoV15} &X&-&- & 74.5\\ 
			\arrayrulecolor{cwblue1} \midrule
			Depth Fusion \cite{Zhu2015} &-&-&X & 88.8\\ 
			MMMP  \cite{Shahroudy20163}  &X&-&X & 91.3\\ 
			DL-GSGC \cite{Luo2013} &X&-&X & 95.0\\ 
			DSSCA - SSLM \cite{Shahroudy20162}  &-&X&X & 97.5\\ 
			\arrayrulecolor{cwblue1} \midrule
			\textbf{Ours (Pose only)}      &X&-&- & 74.6       \\ 
			\textbf{Ours (RGB only)}      &-&X&- & 75.3       \\ 
			\textbf{Ours (Pose + RGB)} &X&X&- & 90.0 \\ 
			\arrayrulecolor{cwblue1} \bottomrule
		\end{tabular} 
	\end{center}
	\caption{Results on MSR Daily Activity 3D dataset (accuracies in \%)}
		\label{MSR}
\end{table}

\begin{table}[t]
    \begin{center}
        \begin{tabular}{cccc}
            \arrayrulecolor{cwblue1} \toprule
            Methods & CS & CV & Avg \\ 
            \arrayrulecolor{cwblue1} \toprule
            Random joint order   &    75.5    &  83.2& 79.4\\
            Topological order w/o double entries   &   76.2       &  83.9& 80.0\\
            Topological order    &   77.1      & 84.5 & 80.8 \\
            \arrayrulecolor{cwblue1} \bottomrule
        \end{tabular}
    \end{center}
    \caption{Results on NTU: pose only, effect of joint ordering.}
    \label{topo_results}
\end{table}

\begin{table}[t]
    \begin{center}
        \begin{tabular}{ccccc}
            \arrayrulecolor{cwblue1} \toprule
            Methods & \multicolumn{1}{c}{Attention} & CS& CV & Avg \\ 
             & {\footnotesize Conditional to pose} & \\
            \arrayrulecolor{cwblue1} \toprule
            RGB only &- & 66.5 &   72.0    & 69.3    \\ 
            RGB only &  X&  75.6   &  80.5 & 78.1 \\ 
            Multi-modal &- & 83.9 &   90.0    & 87.0    \\ 
           Multi-modal &    X&  84.8  &  90.6 & 87.7 \\ 
                        
            \arrayrulecolor{cwblue1} \bottomrule 
        \end{tabular}
    \end{center}
    \caption{Results on NTU: conditioning the attention mechanism on pose (RGB only, accuracies in \%).}
    \label{adding_pose}
\end{table}

\begin{table*}[t]
    \begin{center}
        \begin{tabular}{cccccccccc}
            \arrayrulecolor{cwblue1} \toprule
            & Methods    & {\footnotesize Pose} & {\footnotesize RGB} & 
            \multicolumn{3}{c}{Attention} & CS& CV & Avg \\
            &   & & & {\footnotesize Spatial} & {\footnotesize Temporal}  & {\footnotesize Pose}& \\
            \arrayrulecolor{cwblue1} \toprule
            A & Pose only                    & X & - & - &-&-& 77.1         & 84.5     & 80.8   \\
            \arrayrulecolor{cwblue1} \midrule
            B & RGB only, no attention (sum of features)        & - & X & - & - &-&  61.5         & 65.9       &  63.7 \\ 
            C & RGB only, no attention (concat of features)        & - & X & - & - &-& 63.2         & 67.2      &  65,2  \\ 
            \arrayrulecolor{cwblue1} \midrule
            E & RGB only + spatial attention      &$\circ$&X&X&-&X& 67.4 &   71.2    &  69.3 \\ 
            G & RGB only + spatio-temporal attention     &$\circ$&X&X&X&X&75.6   &  80.5 & 78.1\\ 
            \arrayrulecolor{cwblue1} \midrule
            H & Multi-modal, no attention (A+B)      & X & X & - & - & -&83.0         & 88.5       & 85.3 \\ 
            I & Multi-modal, spatial attention  (A+E)       & X & X & X & - &X&    84.1    &   90.0  &   87.1   \\ 
            K & Multi-modal, spatio-temporal attention  (A+G) & X & X & X & X       &X&   84.8      &   90.6 &  87.7 \\ 
            \arrayrulecolor{cwblue1} \bottomrule 
        \end{tabular}
    \end{center}
    \caption{Results on NTU: effect of attention. $\circ$ means that pose is only used for the attention mechanism.}
        \label{effect_attention}
\end{table*}

\myparagraph{Implementation details}
Following \cite{Shahroudy2016}, we cut videos into sub sequences of 20 frames and sample sub-sequences. During training a single sub-sequence is sampled, during testing 10 sub-sequences and logits are averaged. We apply a normalization step on the joint coordinates by translating them to a body centered coordinate system with the ``middle of the spine'' joint as the origin (gray joint in figure \ref{fig:toposkel}). 
If only one subject is present in a frame, we set the coordinates of the second subject to zero.
We crop sub images of static size on the positions of the hand joints ($50{\times}50$ for NTU, $100{\times}100$ for MSR and SBU). Cropped images are then resized to $299{\times}299$ and fed into the Inception model. 

Training is done using the Adam Optimizer \cite{AdamOptimization2015} with an initial learning rate of 0.0001. We use minibatches of size 64 and dropout with a probability of 0.5. Following \cite{Shahroudy2016}, we sample 5\% of the initial training set as a validation set, which is used for hyper-parameter optimization and for early stopping. All hyperparameters have been optimized on the validation sets of the respective datasets.

When transferring knowledge from NTU to MSR and SBU, the target networks were initialized with models pre-trained on NTU. Skeleton definitions are different and were adapted. All layers were finetuned on the smaller datasets with an initial learning rate 10 times smaller then the learning rate for pre-training.

\myparagraph{Comparisons to the state-of-the-art}
We show comparisons of our models to the state-of-the-art methods in table \ref{NTU}, table \ref{SBU} and table \ref{MSR}, respectively. We achieve state of the art performance on the NTU dataset with the pose stream alone or with the full model fusing both streams. On the SBU dataset, we obtain state of the art performance with the full model, on the MSR dataset we are close.

As mentioned, the reported performances on the NTU and MSR datasets include a knowledge transfer from the NTU dataset. Results on MSR show the difficulty of training a fully learned representation on a tiny dataset. We outperform all methods in the first group of table \ref{MSR}, which correspond to hand-crafted approaches.

We conducted extensive ablation studies to understand the impact of our design choices.

\myparagraph{Joint ordering}
The joint ordering in the input tensor $\bX$ has an effect on performance, as shown in table \ref{topo_results}.
Following the topological order described in section \ref{subsec:skeletonnet} gains ${>}1$ percentage point on the NTU dataset w.r.t. random joint order, which confirms the interest of a meaningful hierarchical representation. As anticipated, keeping the redundant double joint entries in the tensors gives an advantange, although it increases the amount of trainable parameters.

\myparagraph{The effect of the attention mechanism}
\noindent
The attention mechanism on RGB data has a significant impact in term of performance as shown in table \ref{effect_attention}. We compare it to baseline summing (B) or concatenating (C) features. In these cases, hyper-parametres where optimized for these meta-architectures. The performance margin is particularly high in the case of the single stream RGB model (methods E and G). In the case of the multi-modal (two-stream) models, the advantage of attention is still high but not as high as for RGB alone. A part of the gain of the attention process seems to be complementary to the information in the pose stream, and it cannot be excluded that in the one stream setting a (small) part of the pose information is translated into direct cues for discrimination through an innovative (but admittedly not originally planned) use of the attention mechanism. However, the gain is still significant, with $\sim$2.5 percentage points compared to the baseline.

Figure \ref{fig:attentionexample} shows an example of the effect of the spatial attention process: during the activity of \emph{Putting an object into the pocket of somebody}, the attention shifts to the ``putting'' hand at the point where the object is actually put.





\myparagraph{Pose-conditioned attention mechanism}
Making the spatial attention model conditional to the pose features $\bs$ is confirmed to be a key design choice, as can be seen in table \ref{adding_pose}. In the multi-modal setting, a full point is gained, ${>}$12 points in the RGB only case.

\myparagraph{Runtime}
For a sub-squence of 20 frames, we get the following runtimes for a single Titan-X (Maxwell) GPU and an i7-5930 CPU: A full prediction from features takes $1.4$ms including pose feature extraction. This does not include RGB pre-processing, which takes additional 1sec (loading Full-HD video, cropping sub-windows and extracting Inception features). Classification can thus be done close to real-time. Fully training one model (w/o Inception) takes $\sim$4h on a Titan-X GPU. Hyper-parameters have been optimized on a computing cluster with 12 Titan-X GPUs.
The proposed model has been implemented in Tensorflow.

\section{Conclusion}
\label{sec:conclusion}

\noindent
We propose a general method for dealing with pose and RGB video data for human action recognition.
A convolutional network on pose data processes specifically organized input tensors. A soft-attention mechanisms crops on hand joints allows the model to collect relevant features on hand shape and on manipulated objects. Adaptive temporal pooling further increases performance. Our method shows state-of-the-art results on several benchmarks and, up to our knowledge, is the first method performing attention on pose and RGB and the first method performing knowledge transfer in human action recognition.


{\small
\bibliographystyle{ieee}
\bibliography{egbibshort}
}

\end{document}